# ViS-HuD: Using Visual Saliency to Improve Human Detection with Convolutional Neural Networks

Vandit Gajjar[1*¥]   Yash Khandhediya[1,2*¥£]   Ayesha Gurnani[1€]   Viraj Mavani[1€]
Mehul S. Raval[3]

[1]Computer Vision Group, L. D. College of Engineering
[2]Dosepack LLC, Meditab Software Inc.
[3]School of Engineering and Applied Science (SEAS), Ahmedabad University
{gajjar.vandit.381, gurnani.ayesha.52, mavani.viraj.604}@ldce.ac.in,
yashk@dosepack.com, mehul.raval@ahduni.edu.in

## Abstract

*The paper presents a technique to improve human detection in still images using deep learning. Our novel method, ViS-HuD, computes visual saliency map from the image. Then the input image is multiplied by the map and product is fed to the Convolutional Neural Network (CNN) which detects humans in the image. A visual saliency map is generated using ML-Net and human detection is carried out using DetectNet. ML-Net is pre-trained on SALICON while, DetectNet is pre-trained on ImageNet database for visual saliency detection and image classification respectively. The CNNs of ViS-HuD were trained on two challenging databases - Penn Fudan and TUD-Brussels Benchmark. Experimental results demonstrate that the proposed method achieves state-of-the-art performance on Penn Fudan Dataset with 91.4% human detection accuracy and it achieves average miss-rate of 53% on the TUD-Brussels benchmark.*

## 1. Introduction

Human detection is an important topic in the field of computer vision [3, 4, 5, 6, 7]. The challenge for the detection has to overcome variations in human pose, light conditions, cluttered background, viewpoint variations and low resolution. Some challenging examples are shown in Figure 1. Moreover, a detection algorithm should be robust to occlusion and cluttered background in the frame.

Many human detectors [6, 8, 9, 10] have been developed to address these challenges. They extract features, such as Histogram of Oriented Gradients (HOG) [3], Haar-like descriptors [5], or their combinations [6, 11], from images and then apply classifiers as boosting [12], Support Vector Machines (SVM) [3], and structure SVM [6] detect a human(s) in a frame. Introduced by Dalal *et al.* [3], HOG is most popular amongst other approaches for human detection. HOG delivered significant improvements and therefore it is an important baseline feature. In order to cope with partial occlusions, Wang *et al.* [6] combined HOG features with the local binary pattern (LBP). Mainly, the model exploits weak representations based on features.

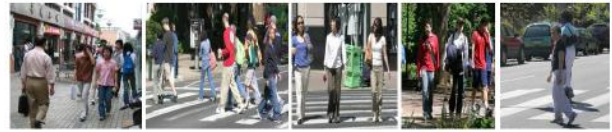

Figure 1: Sample images with cluttered background and strong occlusions.

Currently, CNN has outperformed handcrafted feature-based classifiers. Many attempts have been made to predict several objects in the close range [41, 42, 43]. Currently, object detectors operate either by scanning the image in sliding window [10, 44] or by using the mechanism described in [45, 46]. Deformable part-based models (DPM) [8, 9, 10, 13, 14] are proposed to handle human pose variations. In order to handle more complex and larger variations, a learned mixture of each body part [14, 15] is used. E.g., Poselets. [15] are learned by clustering pose annotations and region appearances.

Over the last decade, deep learning has been applied to human detection with promising results [4, 16, 17, 18]. Instead of using handcrafted features, it automatically learns features in an unsupervised or supervised learning e.g., Restricted Boltzmann Machine (RBM), and discriminative RBM. The network is often stacked in many layers to map the raw data into higher-level representations [4, 19]. Then, the entire network is fine-tuned with given label information and the output of top-layer is often adopted as features to train a classifier. However, the hierarchical representations learned by deep model do not have semantic meanings e.g., the body parts; upper and lower, as in previous hierarchical DPM [8, 9, 10, 13, 14, 20]. Ouyang *et al.* [18] and Girshick *et al.* [21] extend DPM to a deep model by learning feature representations and jointly optimizing the key components of DPM. However, the models did not suppress the influence of cluttered background. The above approaches work well for images with fewer non-occluding objects. On a closer analysis these methods fail to capture visual features of a human body during strong occlusions. As a solution, we propose to use visual saliency to improve human detection in the cluttered background or strong occlusions.

## 1.1. Visual Salient Object Detection

Visual saliency detection is closely related to selective processing in the human visual system. It aims at highlighting visually salient regions or objects in an image. It is fundamentally an intensity map where higher intensity signifies regions which draw human attention. Over the last decade, many improvements have been witnessed in visual saliency detection. It is computed using different algorithms and methods, after years of research in the field of cognitive science. Many supervised and unsupervised visual saliency detection methods have been proposed under several theoretical models [33, 34, 35]. Most unsupervised algorithms and methods are based on low-level features and perform saliency detection on the individual image. Itti *et al.* [34] proposed a saliency model which linearly combines image features including color, intensity, and orientation over various scales to detect local prominence. Still, this method tends to enlighten the salient pixels and loses object information. Zhu *et al.* [33] propose a background measurement to characterize the spatial layout of image regions. Cheng *et al.* [36] address saliency detection based on the global region contrast, which simultaneously considers the spatial coherence across the regions and the global contrast over the image.

The methods that consider only local contexts tend to detect high-frequency content and suppress the region inside salient objects. Achanta *et al.* [37] estimate visual saliency by computing the color difference between each pixel with respect to its mean. The work proposed by Liu *et al.* [38] uses both local and global set of features, which are integrated by a random field to generate a saliency map. Yan *et al.* [39] proposed a multi-layer approach to analyze high contrast regions. However, most method discussed above integrates hand-crafted features to create the final saliency map.

Recently, researchers uses visually salient features and CNN for classification or detection problems, e.g., Uddin et al. [47], Liu et al. [48], Happy et al. [49], Mavani *et al.* [58], Gurnani *et al.* [57] for facial expression classification, Gajjar et al. [50], Wang et al. [51], Aguilar et al. [52] for human detection and tracking in video surveillance, Tong et al. [53] for face detection. Visual saliency has also been exploited in high-level vision tasks, e.g., object detection [16], person re-identification [22, 23].

Visual saliency detection is still an open problem considering heavy occlusion or cluttered background as presented in a complex benchmark like the MIT Saliency Benchmark [28]. Our primary goal in this work is to improve the human detection performance with such difficult background. Towards this goal we use the Deep Multi-Level Network (ML-Net) [1] which has outperformed many other models on the SALICON Dataset [40] and it also performs better on the MIT Saliency Benchmark.

Our novel method uses visual saliency to detect humans under heavy occlusion or cluttered background. We use a direct feature learning process by computing visual saliency maps of the input image and then we multiply both; the image and its visually salient map. These results in the more efficient feature learning by subsequent CNN; the DetectNet resulting into higher detection accuracy in spite of occlusions.

Overall, this paper makes the following contributions:

- We propose a novel method to learn features after computing visual saliency in order to accurately localize humans in spite of heavy occlusion and cluttered background.

- We showcase state-of-the-art results on the challenging Penn-Fudan dataset [29] and achieve a competitive result on Tud-Brussels benchmark [30].

Rest of the paper is as follows: Section 2 describes the DetectNet architecture and a Visual Saliency Model using ML-Net. Section 3 discusses the proposed method ViS-HuD with the pre-processing steps, data augmentation, and implementation details. Dataset, preparation for training, validation and testing image-set, experiments and results are shown in Section 4. Section 5 concludes the paper.

## 2. Preliminaries

### 2.1. DetectNet

DetectNet is a standard model and uses NVIDIA – DIGITS with Caffe deep learning framework as back-end. DetectNet training data samples are large images that contain multiple objects. It has fixed 3-dimensional labeling format which enables to ingest images of different size with a change in number of objects. Figure 2 shows the process of annotated training images for training DetectNet. Each grid square is labeled with two key pieces of information: 1) the class of human presence in the grid square; 2) the pixel coordinates for the corners of the bounding box around the human.

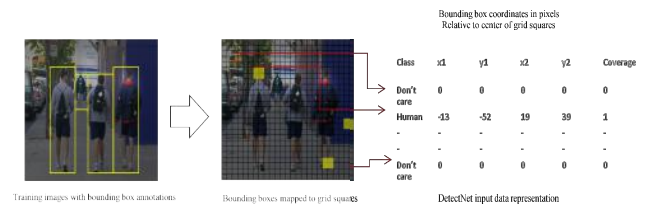

Figure 2: Input data representation for DetectNet.

The coordinates are relative to the center of the grid square. When a human is absent in the grid square, a don't care class is used to maintain fixed size of the data. A coverage value of 0 or 1 is also provided to indicate whether a human is present within the grid square. In the case of multiple humans in the same grid square, DetectNet selects the human that occupies the most pixels within grid squares.

DetectNet architecture points out three very important processes during training.

- A fully-convolutional network (FCN) performs feature extraction and prediction of human class and bounding boxes per grid square.
- Loss functions simultaneously measure the error in the two tasks of predicting the human coverage and human bounding box corners per grid square.
- A clustering function produces the final set of predicted bounding boxes during testing.

The FCN sub-network of DetectNet has the same structure as GoogLeNet [2] except without the data input layers, final pooling layer and output layers. We chose DetectNet, as it can be initialized using a pre-trained GoogLeNet model, thereby reducing training time and improving final model accuracy. Furthermore, the network can accept input images with varying sizes and effectively applies a CNN in a stride sliding window fashion.

## 2.2. The Visual Saliency Model: ML – Net [1]

We used the Deep Multi-Layer ML-Network for visual saliency prediction. A CNN is used to compute low and high-level features from the input image. Extracted features maps are then fed to an Encoding network, which learns a feature weighting function to generate visual saliency-specific feature maps. The main purpose of using a visual saliency model is to propose possible regions where humans might be present in the input image.

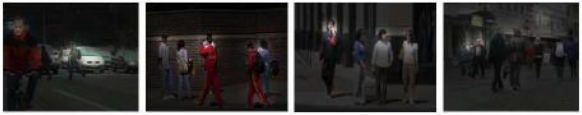

Figure 3: Visual salience image samples after passing from ML-Net and multiplied by the input image.

Visual saliency maps would indicate higher intensity, where there are humans in the image. All the training images of Penn-Fudan and TUD-Brussels are passed from the ML-Net and then multiplied with the corresponding input images. Examples of the salience-windowed images are given in Figure 3. The Deep Multi-Layer Network by Cornia *et al.* described in [1] outperforms all other models, and on a saliency prediction metrics achieves normalized scan path saliency (NSS) score on Correlation Coefficient (CC) of 0.74, AUC shuffled of 0.76, and AUC Judd of 0.86.

## 3. The Proposed Method

The detailed representation of our proposed method (ViS-HuD) is shown in Figure 4. The process starts by passing training images of both the datasets images into ML-Net [1] to compute the visual saliency maps. The network, which is based on a CNN, is used to extract the necessary feature. After computing visual saliency, we scale saliency map by a scalar (0.8) and multiply it to the input image to generate *multiplied visual salient image* (MVSI). Using labels and MVSI, we train the DetectNet. The B-Box regressor predicts, B-Box corner per grid square. The total training loss is the weighted summation of the following losses:

- $L_1$: Loss of the coverage maps estimated by the network and ground truth.

$$\frac{1}{2N}\sum_{i=1}^{N} | coverage_i^t - coverage_i^p |^2 \ (1)$$

The coverage map extracted from annotated ground truth is coverage$^t$ and the predicted coverage map, while denoting the batch size is coverage$^p$.

- $L_2$: Loss between the true and predicted corners of the B-Box box for the human covered by each grid square.

$$\frac{1}{2N}\sum_{i=1}^{N}[\ |x_1^t - x_1^p| + |y_1^t - y_1^p| + |x_2^t - x_2^p| + |y_2^t - y_2^p|\ ] \ (2)$$

Where, $(x_1^t, y_1^t, x_2^t, y_2^t)$ are the ground-truth B-Box co-ordinates, while $(x_1^p, y_1^p, x_2^p, y_2^p)$ are the predicted B-Box coordinates.

During the testing stage, we threshold the coverage map obtained after passing image through the FCN network, and use the B-Box regressor to predict the corners. Since multiple bound-boxes are generated, we finally cluster them to refine the predictions.

### 3.1. Preprocessing

In our method CNN requires fixed size for training, so we rescale images from both the dataset to a size of 512×512.

### 3.2. Data Augmentation

It is common in computer vision to augment the training samples from original data to increase the robustness. E.g., the Penn-Fudan Dataset contains 170 images with 345 humans. These are not sufficient in number rendering the dataset prone to overfitting. We have used different data augmentation techniques to prevent overfitting. First, we flipped all images horizontally. Then, each image is rotated and horizontally flipped with angles $\{-7°, -5°, -3°, 3°, 5°, 7°\}$. This makes the model robust against the low rotational changes in the input image. All the variants result into dataset which is 13 times bigger than original. Figure 5 shows augmented samples for Penn-Fudan.

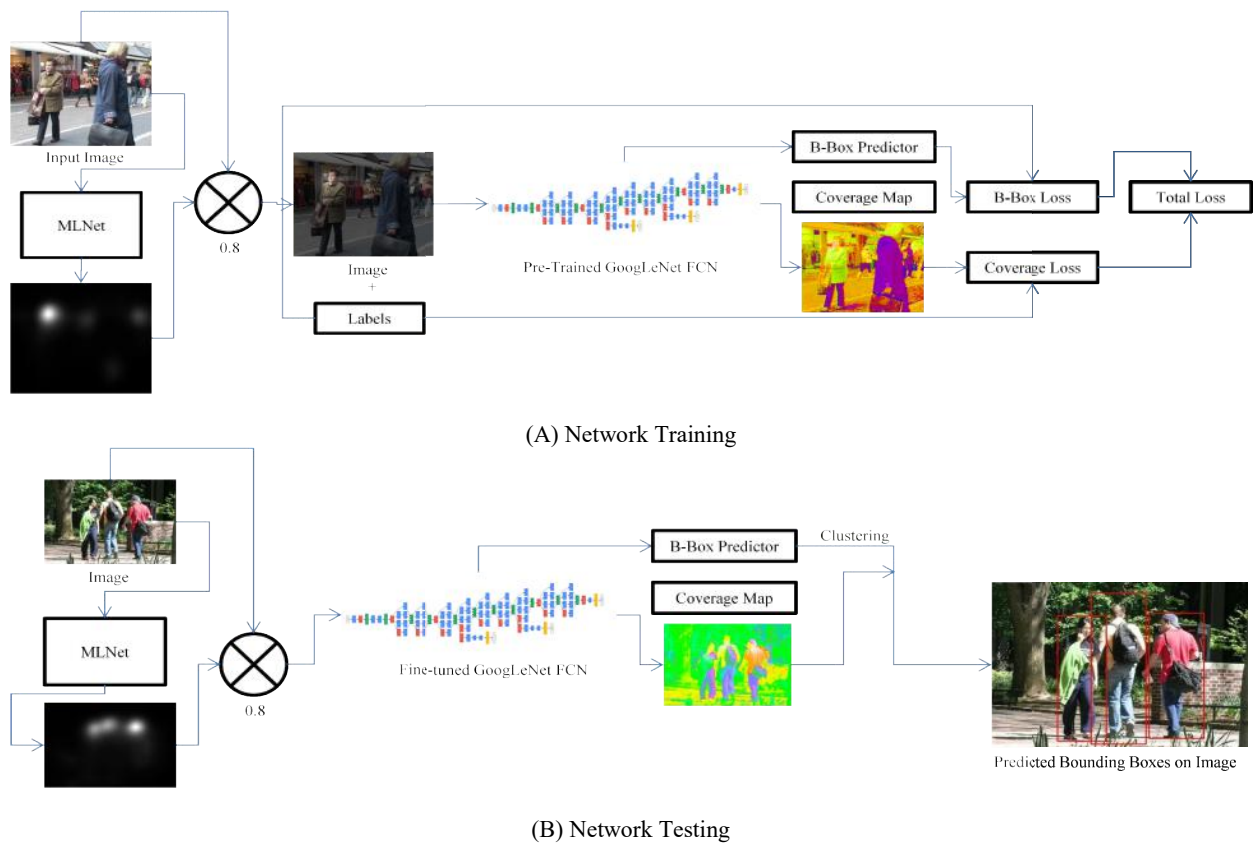

Figure 4: Our method for detection of humans using DetectNet. (A) Using the salient images with annotation, the FCN generates the coverage map and B-Box co-ordinates. Total training loss is the weighted sum of coverage and B-Box loss. (B) At testing time, coverage map and B-Box are generated from the FCN.

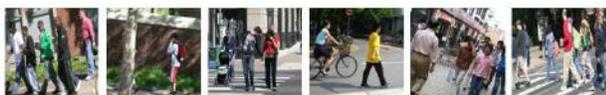

Figure 5: Augmented samples for Penn-Fudan Dataset.

### 3.3. FCN Training in DetectNet

Figure 4 illustrates training and testing architecture of DetectNet [54]. During training, the visual salient images with dimension 512 x 512 are labeled with data format required for DetectNet; i.e., human class, B-box coordinates, and coverage value. The training data is given to fully convolutional network (FCN) predicting coverage map for human class. The detailed representation of FCN network is shown in Figure 6.

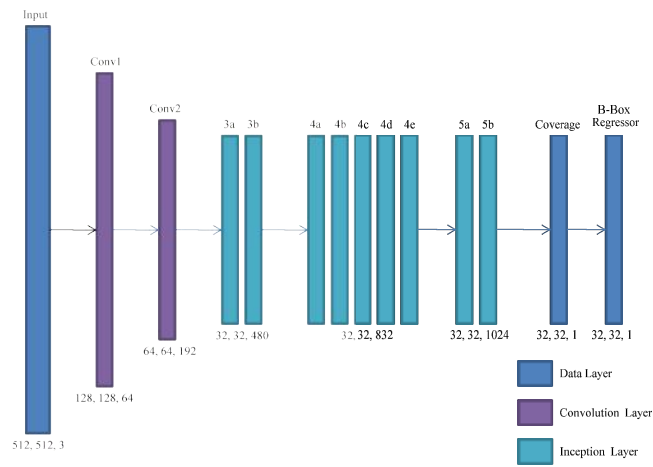

Figure 6: The FCN used in the DetectNet.

Table 1: Overview of training, validation and testing image set.

| | Training | | | Validation | | | Testing | | |
|---|---|---|---|---|---|---|---|---|---|
| | Human Images | Negative Images | Positive Images | Human Images | Negative Images | Positive Images | Human Images | Negative Images | Positive Images |
| Penn-Fudan | 1000 | - | - | 350 | - | - | 1030 | - | - |
| TUD-Brussels | 1500 | 218 | 1092 | 276 | - | - | 1498 | - | 508 |
| Subtotal | 2500 | 218 | 1092 | 626 | - | - | 2528 | - | 508 |
| Total | Training Images – 3810 | | | Validation Images – 1252 | | | Testing Images - 3036 | | |

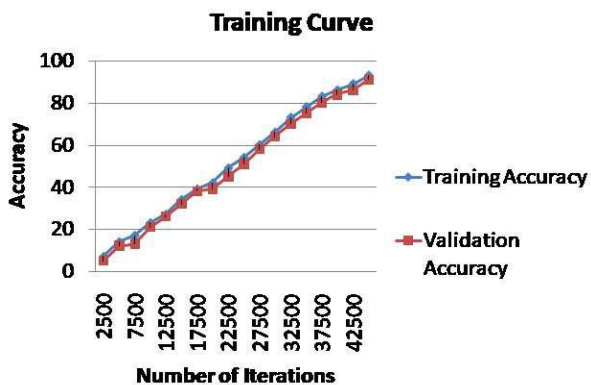

Figure 7: Performance evaluation plot of Network on the training and validation image set, after 45000 iterations.

We have changed the number of class to only one; i.e., with a human. The Convolution layer includes Convolution, ReLU activation, and Pooling. As described in GoogLeNet [2], the inception layer includes the same module. For our experiments, we use pre-trained weights on ImageNet to initialize the FCN network.

We have used the workstation of Intel Xeon core processor accelerated by Quadro K5200. The model is trained for 90 epochs with each of 500 iterations. Figure 7 shows the training curve. We then train the model by minimizing the cross entropy loss function. Using Adam optimizer, we train the FCN through stochastic gradient descent with the learning rate of 0.0001. After 60 epochs, for every 10 epochs the learning rate is dividing with 10. We have observed little over-fitting during training the model thus; we increased the dropout to 0.5. The overall training process has taken more than one-day to complete.

## 4. Experiments and Results

We evaluate our method on two public databases: Penn-Fudan Dataset [29] and TUD-Brussels [30].

### 4.1. Datasets

The Penn-Fudan Dataset consist 170 images with 345 labeled pedestrians. The images are taken from the urban street and labeled pedestrians fall into 390 x 180 pixels.

TUD–Brussels data is conceivably most popular dataset for human detection and it comes with pre-defined subsets for training and testing. The training dataset consist 2810 images (1776 annotated human labels, 1092 positive sample, and 218 negative samples). The testing dataset has 2006 images including 508 positive samples.

#### 4.1.1 Preparation of Training and Testing Data

We prepare training, validation and testing split as per given table. We took care for splitting the data as the training and testing images do not overlap with each other. Therefore, using both datasets, total 3810 images are available for the training set, 1252 images for validation set and 3036 images (1030 images of Penn-Fudan and 2006 images of TUD-Brussels) are presented for the testing set. Table 1 shows the overview of training, validation, and testing image set.

### 4.2. Evaluation Metrics

For Penn-Fudan dataset, we evaluate the detection performance on the test set comparing detection results with the available ground truth and for TUD-Brussels, we follow evaluation criteria proposed in [55], where the log-average miss rate is calculated by averaging the miss rate at nine False Positive per Image rates, which are evenly spaced in log-space in the range from $10^{-2}$ to $10^{0}$.

We compare with the best-performing methods; i.e., HOG, Adaboost, SVM-HOG, and Adaboost-HOG for Penn-Fudan Dataset, while for TUD-Brussels benchmark, we compare our performance using Viola Jones (VJ), HOG, Aggregated Channel Features (ACF), Multi-Feature + Motion, Roerei, Integral Channel Features Detector (ChnFtrs), CrossTalk, Macro Feature Layout Selection (MLS), pAUCBoost, Locally Decorrelated Channel Features (LDCF), SpatialPooling, FPDW, FisherBoost, MF + Motion + 2PED, MultiFtr + CSS, ADABoost, SVM-HOG, and Adaboost-HOG methods. These methods detect humans on static images, and do not use external video motion information. We have performed three exhaustive experiments in order to showcase the effectiveness of our method.

### 4.2.1 Cross-Dataset Validation

We performed cross-dataset validation without using our method, i.e., without using Saliency maps. We train DetectNet on one dataset and test on another. The detection accuracy on Penn-Fudan dataset was considerably low with 73.1%, and we achieve a very low average miss-rate of 71% on TUD-Brussels benchmark. The different viewpoint is the main factor affecting the performance as Penn-Fudan dataset have images in mid-field range, while TUD–Brussels benchmark has many far-field images compared to mid-field images. Thus, in order to improve the detection performance, we explored the use of visual saliency.

### 4.2.2 Improving Performance using ViS-HuD

Initially we trained DetectNet (without saliency) on Penn-Fudan and TUD-Brussels dataset. When tested on images from the respective dataset it resulted into accuracy of 81.3% for Penn-Fudan and average miss-rate of 67% for TUD-Brussels dataset.

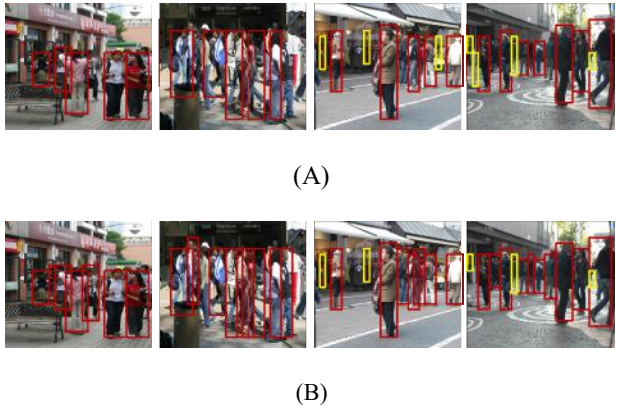

Figure 8: (A) Some of the test objects are not detected and miss-rate was comparatively high when network was trained without using visual saliency (B) Correct detection by the trained network using visual saliency on same test images. Red shows the True Positive; while yellow shows the False Positive. First two images are of Penn Fudan dataset, while other two are of TUD-Brussels.

Secondly, using our method, as shown in Figure 4, resulted into detection accuracy of 85.7% Penn-Fudan test set and average miss-rate of 57.9% on TUD-Brussels test set. One can note significant improvement due to use of the saliency maps as direct feature learning process helps the network architecture. Figure 8 shows the results on test images; without saliency and with our method. The detection accuracy improves with the use of saliency map before person detection.

### 4.2.3 Ablation Experiments

To further improve the detection performance, we perform ablation experiments. Using both the dataset's training images; we trained the network and tested the model on test images of the individual dataset. By combining both the datasets, the network learns the features of images in mid and far-field. Thus, our method achieves state-of-the-art detection accuracy with 91.4% (correct predictions for 942 images out of 1030 test images). Table 2 shows the comparison of different approaches.

Figure 9 shows the qualitative results on some of the test images of TUD-Brussels and Penn Fudan dataset which contains heavy occlusion and cluttered background. First two images are of TUD-Brussels and other are of Penn Fudan dataset.

Table 2: Comparison with the different approach performance on Penn-Fudan Dataset.

| Approach | Detection Accuracy | Number of Correct Human Detection |
|---|---|---|
| HOG [3] | 67.76% | 697/1030 |
| Adaboost [61] | 72.23% | 744/1030 |
| SVM-HOG [62] | 78.93% | 813/1030 |
| Adaboost-HOG [62] | 85.14% | 877/1030 |
| R-FCN [59] | 86.31% | 889/1030 |
| YOLO [9] | 88.73% | 914/1030 |
| Fast R-CNN [60] | 90.38% | 931/1030 |
| Ours | **91.4%** | **942/1030** |

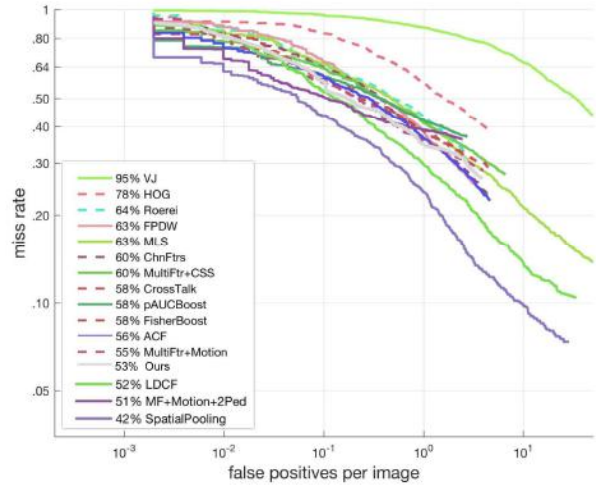

Figure 10: Results of different detection methods on TUD-Brussels dataset using standard evaluation settings.

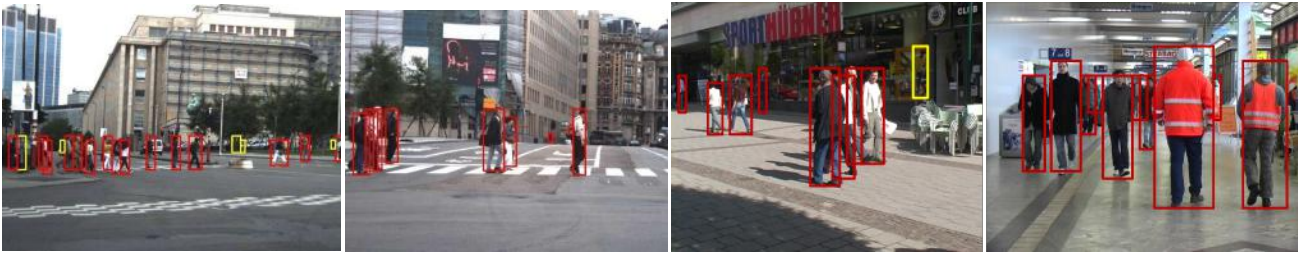

(A) Test images of TUD-Brussels dataset.

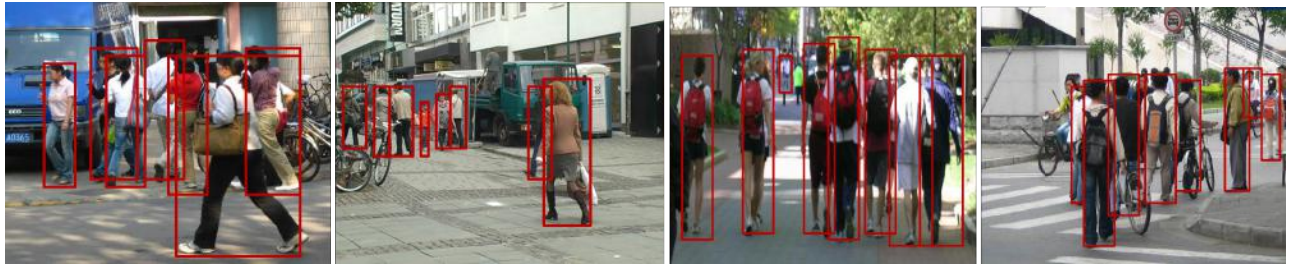

(B) Test images of Penn-Fudan dataset.

Figure 9: Qualitative detection result on some test images of TUD-Brussels and Penn-Fudan dataset which contains heavy occlusions and cluttered background. (A) Red shows the True Positive, while yellow shows the False Positive.

On TUD-Brussels benchmark, we achieve the competitive average miss-rate of 53% on the test set. The results in Figure 10 show that our proposed method outperforms the baseline detection [56] by approximate 7%. It can be observed in general that for all other detection methods, performance drops significantly as the occlusion increases.

## 5. Discussion and Conclusion

We observed that the proposed method converges after 300 iterations per epoch; additional iterations do not gain boost for detection. The limited size of the training images and diversity is the main characteristic of performance. Still, the experimental results on the Penn-Fudan dataset suggest that our detection method is not restricted to specific occlusion patterns or crowd densities. The performance further improved for TUD-Brussels dataset, having more complex and highly occluded scenes.

Our method shows a new direction for training a CNN, that uses multiplied visual salient image and data augmentation. When using two or more challenging benchmarks, an image dimension becomes a very important factor. The detection results can be further improved by using uniform image size and same type of images during pre-training a network.

Using saliency maps we wanted to completely eliminate the background, but sometimes, blurring would create abrupt edges that confuse the filters during the training period. Also, low and mid frequency information remains well preserved when using a visual saliency. Moreover, one can still identify the content of the image, i.e., humans, cloth color from the visually salient image.

Also by giving the multiplied visual salient image to a neural network, it is easy for the network to extract the features. Another promising future extension of this work would be to detect humans on extremely dense scenes and tracking them.

Finally, the proposed method achieved state-of-the-art performance on Penn-Fudan dataset with 91.4% detection accuracy and achieves the competitive result on challenging TUD-Brussels dataset with average miss-rate of 53%.

## Acknowledgments

We are thankful to the anonymous reviewers for their valuable comments due to which the paper was improved. We thank Param Rajpura for the insightful discussions. We would also like to acknowledge the support of NVIDIA Corporation with the donation of the Quadro K5200 GPU used for this research.